# HyperSLICE: HyperBand optimized Spiral for Low-latency Interactive Cardiac Examination

**Submitted to Magnetic Resonance in Medicine (MRM)**

Dr. Olivier Jaubert[1], Dr. Javier Montalt-Tordera[1], Dr. Daniel Knight[1,3], Pr. Simon Arridge[2], Dr. Jennifer Steeden[1], Pr. Vivek Muthurangu[1]

[1.] UCL Centre for Translational Cardiovascular Imaging, University College London, London. WC1N 1EH. United Kingdom

[2.] Department of Computer Science, University College London, London. WC1E 6BT. United Kingdom

[3.] Department of Cardiology, Royal Free London NHS Foundation Trust, London. NW3 2QG. United Kingdom.

Corresponding author:

|  |  |
|---|---|
| Name | Olivier Jaubert |
| Department | Centre for Translational Cardiovascular Imaging |
| Institute | University College London |
| Address | 30 Guilford St, London WC1N 1EH |
| E-mail | o.jaubert@ucl.ac.uk |

*Acknowledgments*: Siemens Healthineers AG (Erlangen, Germany) provided the base code for the interactive sequence. This work used the open source frameworks Gadgetron (v4.1.1) and TensorFlow MRI (v0.18.0).

*Grant Support*: This work was supported by UK Research and Innovation (MR/S032290/1), Heart Research UK (RG2661/17/20) and the British Heart Foundation (NH/18/1/33511, PG/17/6/32797).

*Running Title*: HyperSLICE for low-latency interactive cardiac examination.

*Word Count:* 4521 words

## Abstract

**PURPOSE:** Interactive cardiac magnetic resonance imaging is used for fast scan planning and MR guided interventions. However, the requirement for real-time acquisition and near real-time visualization constrains the achievable spatio-temporal resolution. This study aims to improve interactive imaging resolution through optimization of undersampled spiral sampling and leveraging of deep learning for low-latency reconstruction (deep artifact suppression).

**METHODS:** A variable density spiral trajectory was parametrized and optimized via HyperBand to provide the best candidate trajectory for rapid deep artifact suppression. Training data consisted of 692 breath-held CINEs. The developed interactive sequence was tested in simulations and prospectively in 13 subjects (10 for image evaluation, 2 during catheterization, 1 during exercise).

In the prospective study, the optimized framework -HyperSLICE- was compared to conventional Cartesian real-time, and breath-hold CINE imaging in terms quantitative and qualitative image metrics. Statistical differences were tested using Friedman chi-squared tests with post-hoc Nemenyi test ($p<0.05$).

**RESULTS:** In simulations the NRMSE, pSNR, SSIM and LAPE were all statistically significantly higher using optimized spiral compared to radial and uniform spiral sampling, particularly after scan plan changes (SSIM: 0.71 vs 0.45 and 0.43).

Prospectively, HyperSLICE enabled a higher spatial and temporal resolution than conventional Cartesian real-time imaging. The pipeline was demonstrated in patients during catheter pull back showing sufficiently fast reconstruction for interactive imaging.

**CONCLUSION:** HyperSLICE enables high spatial and temporal resolution interactive imaging. Optimizing the spiral sampling enabled better overall image quality and superior handling of image transitions compared to radial and uniform spiral trajectories.

# Introduction

Interactive cardiovascular magnetic resonance (MR) imaging combines real-time image acquisition, on-the-fly reconstruction and interactive scan-plane control. It is a vital requirement for MR guided cardiac interventions and is increasingly used for image planning, particularly in patients with complex anatomy. Unfortunately, conventional Cartesian interactive real-time imaging[1] is limited by relatively low spatial and temporal resolutions, which restricts potential applications.

State-of-the-art approaches that combine k-space undersampling, efficient k-space filling and iterative reconstructions such as compressed sensing (CS) have enabled high spatial and temporal resolution real-time imaging[2,3]. However, most of these algorithms are not applicable to interactive imaging due to their incompatibility with on-the-fly reconstruction. Nevertheless, newer iterative approaches like Nonlinear Inverse Reconstruction have been able to perform near real-time reconstruction of highly undersampled interactive radial data iteratively[4,5]. This method has been successfully applied to catheter guidance[6], but requires high-end hardware, including multiple GPU systems, to reconstruct images with low-latency.

More recently, machine learning (ML) has been leveraged for reconstruction of highly undersampled MRI. In particular, unrolled ML architectures have been shown to outperform iterative methods, winning the most recent MR image reconstruction challenges[7,8]. Unfortunately, ML techniques that have multiple iterations and rely on data consistency are also unsuitable for low-latency reconstruction. Alternatively, ML networks applied as a single pass post-processing step (deep artifact suppression) can have short inference times[9]. We have previously demonstrated deep artifact suppression can be used to perform on-the-fly reconstruction of radially undersampled data using a single mid-range GPU and laptop[10].

However, our previous approach suffered from suboptimal image quality during interactive scan-plane changes.

Therefore, we propose two modifications to remedy this problem: changing from radial to spiral sampling and changing the network architecture. Spiral sampling is more efficient than radials but comes with many more degrees of freedom that must be optimized. Thus, in this study we attempt to 'find' the optimal variable density spiral trajectory, number of spiral interleaves and interleave ordering using a bandit-based approach - HyperBand[11]. The HyperBand approach allows multiple sampling schemes to be tested during deep artefact suppression training. The sampling pattern combined with the accompanying artefact suppression network that produces the best image quality can then be used for subsequent prospective inference.

The second modification was to the network architecture used for deep artefact suppression. Our previous approach used a UNet like architecture with recurrent layers (Recurrent UNet) that can process 2D images on-the-fly, while still benefiting from temporal redundancies in the data. The main problems with recurrent networks are divergence during training and inference, and slow transitions during scan plane changes[10]. Consequently, we chose to use FastDVDnet[12] that has recently shown promising performance for real-time video denoising, especially in terms of inference times and the handling of image transitions.

The aims of this study were to: 1) jointly find the optimal spiral sampling pattern and train an optimum deep artefact suppression network for interactive imaging, 2) evaluate this HyperBand optimized Spiral for Low-latency Interactive Cardiac Examination (HyperSLICE) framework against radial and uniform spiral trajectories in simulations, and compare it to Recurrent UNet reconstruction with a focus on image transitions, 3) quantitively and qualitatively compare images from the optimized framework in a prospective cohort, to reference breath-hold Cartesian, real-time Cartesian, and state-of-the-art reconstructions (CS and unrolled network) of the spiral data, 4) qualitatively demonstrate the feasibility of using this interactive acquisition and on-the-fly reconstruction during right heart catheter pull back.

## Methods

This study was approved by the local research ethics committee (ref. 19/LO/1561), and written consent was obtained in prospective and retrospective cohorts. All data (training and prospective datasets) were acquired in a single center on a 1.5T system (Aera, Siemens Healthineers, Erlangen, Germany).

### *Training Data*

The training dataset used for the ML networks consisted of 692 ECG triggered breath-held Cartesian balanced steady-state free precession (bSSFP) CINE multi-coil raw data. The dataset included seven different orientations: short axis (SAX), four chamber (4CH), three chamber (3CH), two chamber (2CH), right ventricular long axis (RVLA), right ventricular outflow tract (RVOT), and pulmonary artery (PA). Data was collected in a diverse adult patient population (age: 58.7 ±16.0 years, weight: 77.4±18.6 kg, male/female: 56/36) referred for routine cardiovascular MR (including assessment for ischemia, cardiomyopathy and pulmonary hypertension).

The raw data was acquired with 2x undersampling (with nominal matrix size of 224x272 and 44 autocalibration lines) and reconstructed with GRAPPA to recover fully-sampled reference Cartesian multi-coil k-spaces. From this data, both undersampled input and high-quality target output images were created.

For the proposed method, target magnitude images were reconstructed through Fast Fourier Transformation (FFT) of the fully sampled Cartesian multi-coil k-space data and coil combination using root-sum-of-squares (RSS). The resultant ground truth images were scaled between 0 and 1 and the same scaling was applied to the multi-coil k-space data before undersampling. Undersampled input images (gridded images) were obtained from the rescaled multi-coil cartesian images via forward non-uniform FFT of each coil image according to the investigated trajectories (as described below) to obtain multi-coil spiral undersampled data, followed by backward non-uniform FFT and RSS coil combination.

*Network Architecture*

FastDVDnet is a convolutional network architecture proposed for video denoising which conventionally takes five noisy frames as input and outputs the denoised central frame. This network architecture was empirically chosen for its high performance, fast runtimes, and its robustness to motion[12].

A modified FastDVDnet (Figure 1) was implemented using TensorFlow and Keras[13]. Modifications included: 1) reducing latency between acquisition and display by outputting the deep artefact suppressed last frame (as opposed to central frame), and 2) empirically enhancing performance in our specific application with no global residual connection, no batch normalization, replacing the pixel-wise addition with a more classic channel-wise concatenation and replacing the loss by a structural similarity index (SSIM) based loss.

Our modified FastDVDnet network takes five consecutive frames of undersampled coil-combined gridded magnitude data as inputs, and outputs the deep artifact suppressed fifth magnitude image (Figure 1A). The FastDVDnet network consists of four denoising blocks, all of which have the same architecture, shown in Figure 1B, where the three "Denoising Blocks 1" of the first layer share the same weights. The network is then applied in a sliding window fashion to output consecutive 2D frames (excluding first four frames of the series). The source code for training and testing our framework for interactive MRI reconstruction (using natural images only for sharing purposes) is provided online (https://github.com/mrphys/HyperSLICE.git).

## *HyperBand Optimization*

HyperBand is a bandit-based technique that makes minimal assumptions and aims at speeding up the evaluation of many hyperparameter configurations. In this study we hyperparameterized six variables that controlled the variable density spiral interleave shape[14], and the number and order of spiral interleaves. Each spiral interleave is defined by five parameters: a k-space inner radius (parameter 1) with a given sampling density (parameter 2), a k-space outer radius (parameter 3) beyond which is a given sampling density (parameter 4), and a transition type (parameter 5, e.g. linear, Hanning or quadratic) which controls the density transition between the inner and outer radiuses of k-space. Additionally, the ordering of the spiral interleaves (parameter 6) could either be tiny golden angle (47.3º) or linear (number of interleaves/360º).

We defined the range of allowable repetition times (TR: [2.88, 3.7] ms) and maximum time per frame ($T_{acq}$: 55ms). The number of interleaves was then calculated as the rounded down value: $T_{acq}$/TR. The explored range of parameterized values was restricted to relevant values and are indicated in Table 1.

The trajectory optimization process is depicted in Supporting Information Figure S1. In brief, randomly parametrized spiral trajectories are generated and used to create paired undersampled and ground truth images that are used to train the FastDVDnet network. By monitoring validation SSIM it was possible to allocate more resources and continue training for the most promising trajectory and network combinations.

The maximum resource allocated to a given configuration (i.e. maximum number of epochs) was set to 150, and the ratio of discarded-to-kept configurations between each step of the HyperBand algorithm was set to 5. This led to the modified FastDVDnet architecture being trained for 217 different spiral trajectories over a total of 3404 epochs. The configuration which resulted in deep artifact suppressed images with the best validation SSIM was selected for subsequent studies.

A subset of the dataset (40%) was used for HyperBand optimization to reduce overall training times. The data was split 30/10% for train/validation during the HyperBand parameter optimization of the spiral. Once the optimized trajectory was selected, the network was then retrained with the whole dataset with a 75/10/15% split for training/validation/testing. All training was performed using TensorFlow[13] on a Linux Workstation (with NVIDIA TITAN RTX 24GB).

*Simulation Experiments to Compare Sampling and Networks*

The optimal spiral trajectory (as chosen by the HyperBand algorithm), was compared to tiny golden radial sampling[15] (with matching temporal resolution i.e. 17 radial spokes per frame) and a uniform density spiral trajectory (with TR and ordering matching those of the optimal trajectory) in simulation. For these two additional trajectories, the same network architecture and dataset as for the final optimized spiral trajectory network were used for training. In addition, to evaluate the benefit of the FastDVD network over the previously used Recurrent UNet, we also trained Recurrent UNets for all three sampling patterns, using the same training data. Recurrent UNets were warmed up by reconstructing four frames prior to the first reconstructed FastDVDnet frame for fair comparison.

 A test set of 2D+time images, comprising of 15% of the dataset (103 CINEs), were undersampled, gridded and deep artefact suppressed using the corresponding network for each of the three trajectories.

Reconstructed simulated real-time data from the test set were assessed in terms of Normalized Root Mean Squared Error (NRMSE), peak signal-to-noise ratio (PSNR), SSIM and Laplacian energy (LAPE)[16,17]. Metrics were averaged over five consecutive frames of a CINE in all subjects (total of 515 images). Additionally, SSIM mean and standard deviation in the test cases was measured in twelve frames during which a sharp transition between two random CINEs from the test set occured (at frame six) to assess the performance during interactive scan plane changes.

## Prospective Experiments

Prospective Evaluation of Image Quality

Prospective datasets were acquired in 10 patients undergoing routine cardiovascular MR. In each subject, three separate long axis (4CH, 3CH and 2CH orientations) CINE data were acquired with: 1) the optimized spiral trajectory – HyperSLICE – pixel size=1.7x1.7mm$^2$, temporal resolution=55ms, 2) an ECG triggered breath-held Cartesian reference sequence – nominal pixel size=1.5x1.5mm$^2$, temporal resolution=27.3ms (~10s breath-hold), and 3) a real-time Cartesian sequence – pixel size=2.5x2.5mm$^2$, temporal resolution=97ms. Full acquisition parameters in Supporting Information Table S1.

HyperSLICE interactive images were reconstructed in near real-time during scanning using Gadgetron[18] for low-latency communication with an external computer (Linux Workstation with NVIDIA GeForce RTX 3060 12GB, where reconstruction times were recorded). The optimized spiral raw data were also retrospectively reconstructed using: 1) a simple gridded reconstruction (the equivalent to the input to the network), 2) navigator-less spiral SToRM[19], which is a state-of-the-art CS reconstruction and 3) spiral VarNet[20] reconstruction, which is an unrolled ML network architecture including data consistency. The gridded, SToRM and VarNet reconstructions were performed offline using open-source codes[19–21]. VarNet was retrained on the same dataset and same optimized trajectory as the proposed HyperSLICE network. The Cartesian datasets used for comparison were reconstructed on the scanner platform using the scanner software reconstructions (including GRAPPA) (16).

Prospective image quality was assessed using quantitative and qualitative metrics. Edge Sharpness (ES) was evaluated quantitatively by measuring the maximum gradient of the normalized pixel intensities between left ventricular (LV) blood pool and LV myocardium at peak diastole using four line profiles drawn manually[22]. The line profiles were filtered using a Savitzky-Golay filter (window width, 8 pixels; third-order polynomial) to remove contamination with noise pixels. Signal-to-noise Ratio (SNR) was estimated quantitively as the ratio between signal in the LV blood pool and empty lung region for all CINEs at peak diastole. Because the signal is estimated from magnitude images, a correction factor of 0.66 is applied to account for the non-Gaussian noise distribution[23].

Subjective image quality were assessed by two clinical experts (V.M., D.K) in terms of motion depiction, presence of artefacts and sharpness of the endocardial border. HyperSLICE was compared with Cartesian real-time and gated acquisitions. Videos were shown individually in a randomized order and graded according to a 5-point Likert scale (1 = non-diagnostic, 2 = poor, 3 = adequate, 4 = good, 5 = excellent image quality). However, Likert scoring lacks granularity when comparing different reconstructions of the same raw data. Therefore, ranking was used to compare the alternative reconstructions of the HyperSLICE data. Specifically, SToRM, VarNet and HyperSLICE videos were shown simultaneously in a shuffled order and ranked (1=best rank, 2= middle rank, 3=worst rank) to enforce differentiation.

Prospective Applications: Catheterization

Two patients were scanned using HyperSLICE during routine catheter pull back that is usually performed without imaging towards the end of right heart catheterization (RHC). During the acquisition, the operator interactively moved the slice position to track the catheter for visualization of its position and movements. Gridded, SToRM, VarNet and HyperSLICE reconstructions of the same optimized real-time spiral acquisition were qualitatively compared.

Prospective Applications: Flexibility

To address whether HyperSLICE could adapt to a wide range of motion states (i.e. deep breathing) or required changes in FOV or spatial resolutions, two additional qualitative experiments were performed on a healthy subject.

HyperSLICE data were acquired during four conditions to evaluate robustness to motion: 1) breath-hold, 2) free-breathing, 3) mild exercise and 4) peak exercise. Exercise was performed on a supine MR-compatible cycle ergometer (MR Cardiac Ergometer Pedal, Lode, Groningen, the Netherlands). Images were reconstructed using HyperSLICE, VarNet and SToRM.

HyperSLICE data were also acquired during free-breathing with: 1) the original FOV (400x400 $mm^2$), and base resolution (240), 2) higher base resolution (FOV = 400x400 $mm^2$, base resolution = 288), 3) larger FOV (FOV = 450x450$mm^2$, base resolution = 240), 4) higher base resolution and larger FOV (FOV = 450x450$mm^2$, base resolution = 288).

## *Statistical Analysis*

Friedman chi-squared tests were performed for repeated measures non-parametric distributions. When results were deemed significant ($p<0.05$), a post-hoc Nemenyi test was performed to test all pairwise correspondences and statistical significances ($p<0.05$). All statistical tests were performed in Python using scipy[24].

# Results

## *HyperBand Optimization*

HyperBand optimization took ~ 6 days during which 217 different spiral configurations were tested. An overview of the SSIM curves for tested configurations are depicted in Supporting Information Figure S1. The trajectories and gridded images for the trajectories with the three highest validation SSIM are shown in Figure 2 (corresponding trajectory parameters in Supporting Information Table S2). Compared to uniform density spiral sampling, the optimized trajectories are heavily sampled in the center of k-space with high undersampling factors in the outer portion of k-space. This results in spatial blurring and light aliasing rather than the heavy undersampling artefacts observed in radial and uniform spiral undersampled images. It should be noted that the best sampling pattern had linear rather than tiny golden angle ordering.

## *Training Times*

The final reconstruction networks trained using the entire dataset took 2 hours for FastDVDnet, 12 hours for Recurrent UNet and 33 hours for VarNet.

## *Simulation Results*

NRMSE, PSNR, SSIM and LAPE mean and standard deviation over the test dataset (n=103) for radial, uniform density spiral and optimized spiral trajectories and both the FastDVDnet and Recurrent UNet are provided in Table 2. NRMSE, PSNR, SSIM and LAPE were all statistically significantly higher for the optimized spiral trajectory compared to radial and uniform density spiral trajectories for both network architectures. Interestingly results also show that for both networks the radial trajectory generally performed statistically significantly better than the uniform density spiral. For the optimized trajectory, the FastDVDnet network gave images with superior SSIM and LAPE (p<0.05), while the Recurrent UNet network gave images with better NRMSE and PSNR (p<0.05).

Figure 3 shows image quality obtained during scan plane changes for radial, uniform density spiral and optimized spiral trajectories using the FastDVDnet reconstruction with Figure 4.A showing the corresponding drop and recovery of SSIM computed over the whole test set. The drop in image quality during transitions was minimal for the optimized spiral trajectory compared to radial and uniform spiral sampling with an average SSIM of 0.71 immediately after transition (vs 0.45 and 0.43 for radial and uniform spiral trajectories, respectively). It should be noted that compared to the Recurrent UNet, the FastDVDnet showed much faster recovery of image quality after a change in scan plane for all trajectories as shown in Figure 4.B.

## Prospective Results

### Prospective Evaluation of Image Quality

Representative X-Y images and X-T profiles from prospectively acquired data for reference, Cartesian real-time, spiral gridded, SToRM, VarNet and HyperSlice methods are presented in Figure 5. Subjectively, HyperSLICE performed significantly (p<0.05) better than real-time Cartesian interactive imaging for image sharpness and motion depiction, but slightly worse for image artefacts (Table 3.A). However, HyperSLICE did not reach the image quality of reference Cartesian gated images (p<0.05). HyperSLICE also significantly (p<0.05) outranked StoRM and VarNet reconstructions for all subjective categories (sharpness, artefacts and motion – Table 3.B).

Quantitative SNR and edge sharpness values are provided in Table 3, with HyperSLICE having similar SNR to real-time Cartesian interactive images, but significantly greater edge sharpness.

HyperSLICE also performed favorably compared to SToRM and VarNet reconstructions (Table 3B).

Compared to the acquisition time of 55ms, the gridded reconstruction took on average 33 ms, deep artifact suppression 19 ms and other processes 5ms (including scaling and formatting). However, as the pipeline can perform gridding and deep artifact suppression of separate images in parallel, the proposed HyperSLICE reconstruction has a theoretical maximum output of one frame every 33ms and is mainly limited by the k-space gridding.

Prospective Applications: Catheterization

HyperSLICE was demonstrated interactively during catheter pull back in Supporting Information Video S1 and S2. In Supporting Information Video S1, both the interactive interface and operator movements are displayed, demonstrating true interactive imaging during catheter tracking. Supporting Information Video S2 shows only the HyperSLICE images acquired in a second patient for better visualization. Figure 6 and Supporting Information Video S3 shows a comparison of the same real-time data acquired during catheter tracking, reconstructed using gridding, StoRM, VarNet and HyperSLICE. The StoRM reconstruction which seems to slightly blur out the catheter motion compared to the gridded, VarNet and HyperSLICE reconstructions. VarNet and HyperSLICE reconstructions seemed to accurately reconstruct the catheter although they were trained uniquely on non-catheterized data. Compared to the gridded and VarNet images, HyperSLICE improved the image sharpness.

Prospective Applications: Flexibility

Supporting Information Video S4 shows HyperSLICE acquired during breath-hold, free-breathing, mild exercise and peak exercise. Even with major differences in motion states, image quality is good and motion appears natural. This is in comparison to the SToRM reconstruction which shows poorer motion fidelity and VarNet reconstruction which shows lower image quality.

Finally, the HyperSLICE network generalized well to modifications of the FOV size and imaging resolutions, with qualitatively no differences in image quality in all cases (Supporting Information Video S5).

# Discussion

In this study, we designed a variable density spiral imaging framework for interactive imaging. The main findings of this study were: i) HyperBand selected an optimal spiral sampling pattern that was much more heavily sampled in the middle of k-space than the outer portions, ii) in simulations the optimal spiral sampling pattern (and associated deep artefact suppression) provided better image quality than either uniform density spiral or radial sampling, iii) the FastDVDnet architecture provided better handling of transitions than the previously used Recurrent UNet architecture, iv) in-vivo the optimal sampling pattern (and associated FastDVDnet deep artefact suppression) provided higher spatio-temporal resolution compared to conventional Cartesian interactive imaging and superior image quality compared to SToRM and VarNet reconstructions of the same spiral raw data, and v) the optimized framework provided high quality interactive imaging during MR guided cardiac catheterization.

## *HyperBand Optimization*

Our previous study using radial sampling and deep artefact suppression demonstrated that high resolution interactive real-time imaging was possible, however imaging could still be improved. In this study we decided to move to more efficient spiral sampling to further improve image quality, particularly during scan plane transitions. An important aspect of imaging is optimization of the sampling pattern and several studies have used ML to learn the optimal sampling through differentiable parameterization of trajectories[25–27]. However, with differentiable parametrizations it would have not been possible to search for variable numbers of spiral interleaves, different orderings, and different density transitions. Therefore, we used a bandit-based approach that enables more freedom on the choice of trajectories and is more appropriate for our non-differentiable optimization problem. This approach was computationally intensive (6 days of training). It required the training of hundreds of configurations and thousands of epochs which makes it challenging to apply to the other ML approaches such as Recurrent UNet or VarNet which would have significantly longer training times.

Compared to uniform density spiral sampling, the top three trajectories were much more heavily sampled in the center of k-space. This resulted in spatially blurred and lightly aliased images rather than heavily aliased but sharp images as input. Nevertheless, the reconstructed optimized spiral images still outperformed those from the uniform spiral sampling in terms of all metrics including LAPE which assesses image sharpness. This suggests that the superior reconstruction for this application is closer to deep super resolution rather than deep artefact suppression.

It should be noted that all the reconstructed images from the three top sampling patterns were within 0.01 SSIM points, even though the input gridded images had some significant visual differences. This demonstrates that the HyperBand algorithm can search widely in parameter space to provide a variety of good candidates.

Our HyperBand framework was optimized for spiral bSSFP imaging, but optimization could easily be done for other spiral sequences. For example, spiral sampling could be optimized for low flip angle GRE sequence, which are desirable for intervention due to low heating of metallic guidewires. In addition, the low level of assumptions means the same method could easily be optimized for other types of trajectories (e.g. variable density radial, or rosette) with no requirements for developing new parametrizations that are differentiable with respect to the loss.

*Simulations Experiments*

In simulations, the optimized spiral provided superior image quality compared to radial and uniform spiral sampling for both FastDVDnet and Recurrent UNet architectures. This was particularly visible during abrupt scan plane changes, with both uniform spiral and radial sampling demonstrating poor image quality for several frames after the transition. We believe that this is because the optimized spiral gridded images contain significantly less artefact than both the uniform spiral and radial gridded images. This means that the optimized network does not need to rely as heavily on previous frames (compared to radial or uniform spiral networks) leading to much better handling of transitions.

Interestingly, both radial and uniform spiral sampling demonstrated similarly low image quality compared to HyperSLICE, particularly during transitions in synthetic data. This suggests that simply changing to a spiral trajectory is not sufficient for good image quality and some form of trajectory optimization is necessary. We also showed that replacing the Recurrent UNet with the FastDVDnet architecture improved the image quality during transitions (for all sampling patterns). This is an important finding of our study and demonstrates that good image quality requires both optimized spiral sampling and improved network architecture. Another benefit of FastDVDnet is that it provides much faster and more stable training, with no divergence over time at inference. This is vital for robust use during clinical imaging. Nevertheless, more recent video denoising architectures and improved realistic simulations of the input images could potentially provide better image reconstruction quality[28].

## Prospective Experiments

In the in-vivo experiments, we demonstrated that HyperSLICE provided sharper and more motion accurate images than conventional Cartesian interactive imaging, due to the significantly higher spatio-temporal resolution. Improved resolution will be vital for greater use of interactive imaging in interventional MRI, as well as for direct cardiac evaluation as opposed to purely planning.

Although in our study the HyperSLICE reconstruction performed better than SToRM and VarNet reconstructions, the trajectory was optimized specifically for HyperSLICE. The SToRM and VarNet reconstructions could also probably be improved through sampling optimization, but it is still encouraging that HyperSLICE was not significantly worse than these state-of-the-art reconstructions. In addition, it is important to note that SToRM and VarNet reconstructions are incompatible with interactive imaging due to too slow reconstruction times, and additional requirements such as pre-computation of coil sensitivities and regularization over the full time-series. It might appear surprising that the unrolled VarNet architecture underperforms compared to the image based deep artifact suppression HyperSLICE, however VarNet does not exploit any temporal information, reconstructing each 2D frame independently.

It should be noted that spiral acquired images did have more artefacts than conventional Cartesian images. This is unsurprising as the spiral acquisition is more highly accelerated. Furthermore, spiral sampling is more susceptible to trajectory errors which could lead to suboptimal reconstructed images. Additional corrections of trajectory errors through gradient impulse function measurements[29] could further improve final image quality. More generally including realistic sources of errors such as noise, off-resonance and gradient imperfections could lead to better results for the trajectory selection, network training and final reconstructed image quality.

*Study Limitations*

In our study, we did not observe any obvious hallucinations. However, it is a possible risk of our approach, particularly as our network architecture does not include any data consistency. Further optimizations (including data consistency using unrolled optimizations) might increase reconstruction accuracy and further improve image quality[7,8]. However, as with compressed sensing methods, these will be heavily constrained by reconstruction times. Other approaches could be investigated such as including an additional loss in k-space to penalize data inconsistency, whilst still keeping reconstruction times short.

HyperSLICE provides a method for jointly optimizing the acquisition and reconstruction. Once optimized, the framework lacks flexibility in temporal resolution, field of view and imaging resolutions, which might be suboptimal depending on patient size and heart rate. However, when tested prospectively, the reconstruction network was shown to qualitatively adapt well to clinically relevant changes of FOV or base resolution. Further quantitative assessment of the network's generalizability to changes in sequence parameters might be required to provide a suitable range of use for good image quality before widespread adoption.

The current framework was trained and evaluated on data from a single 1.5T scanner. Evaluating whether a single trajectory and network could generalize well over data from multiple centers, multiple field strengths or from multiple scanner models is necessary for larger clinical dissemination. It could also be relevant to optimize trajectories depending on field strengths, scanner hardware and clinical applications.

Finally, our framework was not used to perform actual catheter guidance as this was a proof-of-concept study. Further work is required to assess the specific improvements in clinical workflow and/or clinical relevance of higher spatial and temporal resolutions for interactive imaging applications. Extended clinical trials, proof of superiority and interesting commercial prospects, would all be required to convince suppliers to integrate the framework into their software for true clinical dissemination.

## Conclusion

After optimization, a variable density spiral interactive acquisition and reconstruction framework, HyperSLICE, showed promising performance in terms of both image quality and reconstruction times and was successfully demonstrated during interactive cardiac MR-guided intervention enabling tracking during catheter pull back with higher spatial and temporal resolutions than conventional interactive imaging.

## Data Availability Statement

The source code for training and testing our framework for interactive MRI reconstruction (using natural images only for sharing purposes) is provided online (https://github.com/mrphys/HyperSLICE.git).

# Figures

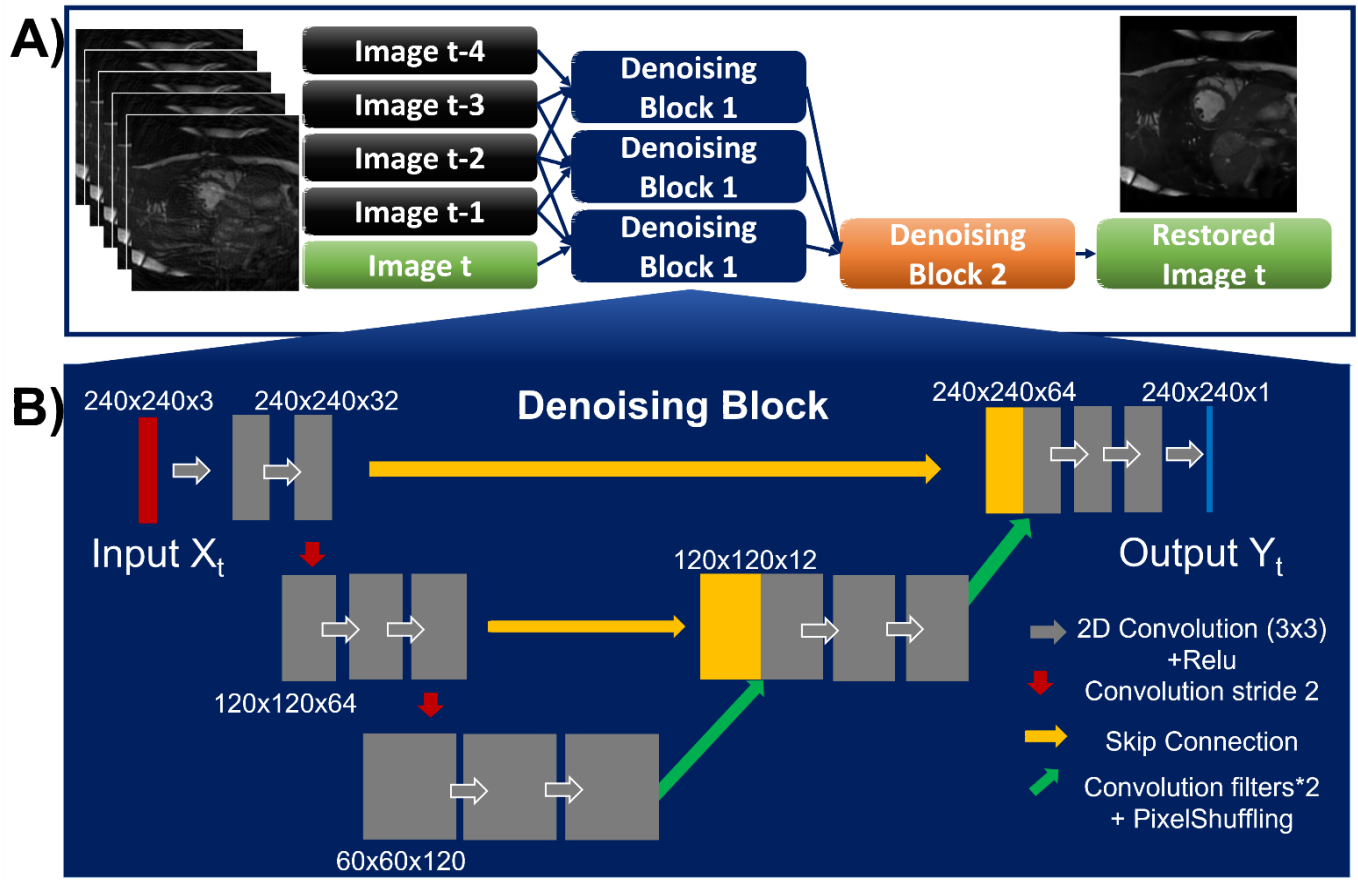

**Figure 1. A)** FastDVDNet modified architecture. Five consecutive undersampled magnitude frames are mapped to the fifth (i.e. latest) ground truth magnitude image for low-latency imaging. All denoising blocks share the same architecture with three input frames and one output frame. The three "Denoising Block 1" in the first layer share the same weights. **B)** The denoising block architecture. Main differences to the original FastDVDnet denoising block include no global residual layer, no batch normalisation layer and concatenation after upsampling.

| Spiral Parameters | Radius delimiting inner k-space | Inner acceleration rate | Radius delimiting outer k-space | Relative acceleration outer/inner | Transition | Ordering | TR (ms) | Temporal resolution (ms) | Inter-leaves | Accel. Rate |
|---|---|---|---|---|---|---|---|---|---|---|
| Range investigated | [0.1:0.3] | [12:24] | [inner:1-inner] | [0.01:0.35] | [linear, Hanning, quadratic] | [linear, tiny golden angle] | [2.88, 3.7] | <55 | / | / |
| Optimized | 0.15 | 16 | 0.56 | 0.07 | Hanning | Linear | 3.67 | 55 | 15 | Variable [1.1,15.0] |
| Uniform | 1 | 92 | 1 | 1 | N/A | Linear | 3.67 | 55 | 15 | 6.1 |

**Table 1.** Range of values tested in HyperBand optimization and resulting optimized spiral imaging parameters as well as matching uniform spiral parameters.

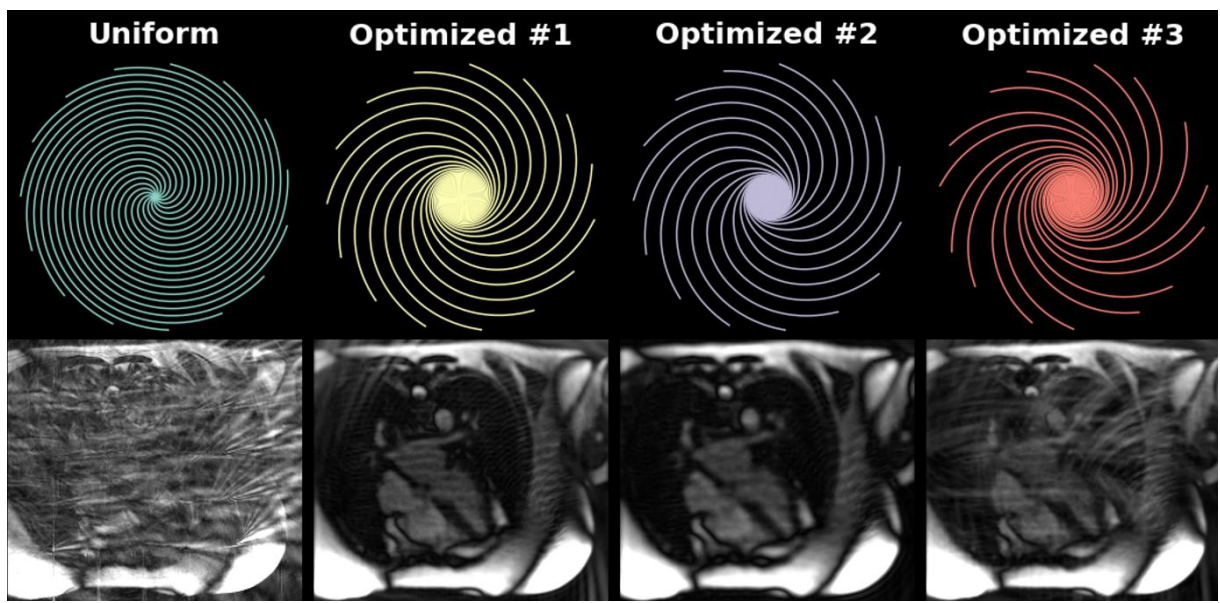

**Figure 2.** Top: Uniform and top three optimized variable density spiral trajectories (first to third highest validation SSIM) obtained from the HyperBand optimization. Bottom: corresponding Gridded images. Corresponding trajectory parameter values and validation SSIM can be found in Supporting Information Figure S2.

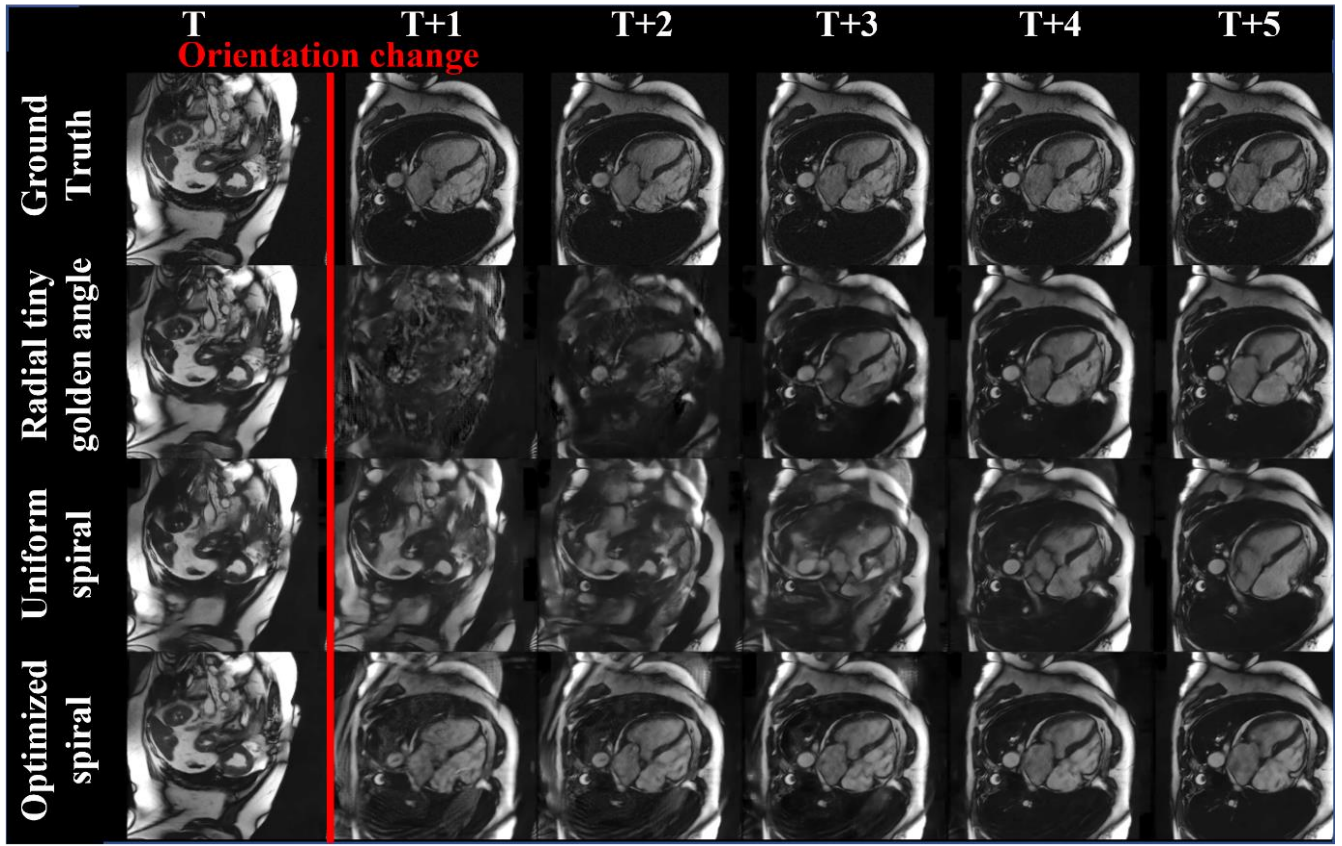

**Figure 3.** Image quality in simulation for six consecutive frames during abrupt scan-plane change from short axis (SAX) to four chambers (4CH). From top to bottom: ground truth and restored radial, uniform spiral and optimized spiral images. Optimized spiral results show better image quality and handling of transitions.

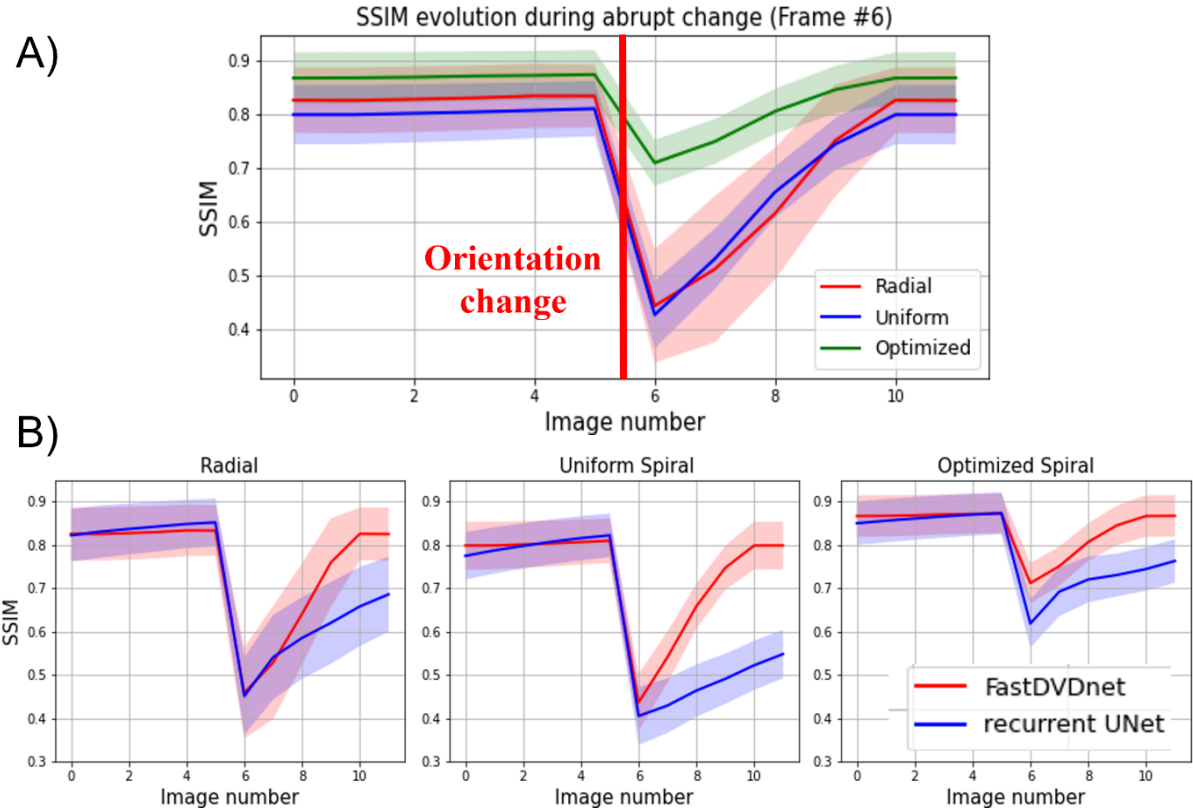

**Figure 4. A)** SSIM through time of Radial, Uniform Spiral and Optimized Spiral FastDVDnet reconstructed images computed over 103 test cases during abrupt scan-plane change performed at frame 6. **B)** SSIM through time comparing FastDVDnet and Recurrent UNet architectures for Radial (left), Uniform Spiral (middle) and Optimized Spiral (right). Frames -4 to 0 were also reconstructed to initialize the Recurrent UNet for fair comparison.

| Test results (n=103) | NRMSE | | PSNR | | SSIM | | LAPE | |
|---|---|---|---|---|---|---|---|---|
| | FastDVDnet | Recurrent UNet | FastDVDnet | Recurrent UNet | FastDVDnet | Recurrent UNet | FastDVDnet | Recurrent UNet |
| Radial | 0.148 ±.033 | 0.141 ±.033 | 30.67 ±2.76 | 31.12 ±2.73 | 0.828 ±.061 | 0.836 ±.059 | 0.487 ±.111 | 0.496 ±.109 |
| Uniform | 0.171 ±.028 | 0.173 ±.029 | 29.33 ±2.43 | 29.25 ±2.46 | 0.802 ±.054 | 0.796 ±.054 | 0.540 ±.099 | 0.564 ±.092 |
| Optimized | 0.127 ±.026 | **0.122 ±.027** | 31.99 ±2.66 | **32.36 ±2.85** | **0.869 ±.047** | 0.860 ±.050 | **0.591 ±.101** | 0.578 ± .099 |

**Table 2.** Mean and standard deviation of NRMSE, PSNR, SSIM and LAPE computed over the test set images (103 slices x5 consecutive frames) for Radial, Uniform and Optimized Spiral reconstructed images using FastDVDnet (Proposed) and Recurrent UNet. Bold values indicate best performing according to metric.

| A) Acquisitions | SNR | Edge Sharpness (mm$^{-1}$) | Subjective Image Sharpness | Subjective Artefacts | Subjective Motion |
|---|---|---|---|---|---|
| Reference Breath-hold | 28.8 ±11.6 | 0.19 ±0.03 | 4.20 ±0.7 | 3.83 ±1.0 | 4.18 ±0.8 |
| Real time Cartesian | 38.3 ±11.7 † | 0.18 ±0.02 | 2.50 ±0.6 † | 3.78 ±0.5 | 2.28 ±0.7 † |
| HyperSLICE | 42.5 ±20.3 † | 0.21 ±0.04 * | 3.67 ±0.5 * | 3.18 ±0.6 †,* | 3.77 ±0.4 * |

| B) Reconstructions | SNR | Edge Sharpness (mm$^{-1}$) | Rank Image Sharpness | Rank Artefacts | Rank Motion |
|---|---|---|---|---|---|
| Spiral gridded | 22.4 ±9.3 | 0.16 ±0.03 | N/A | N/A | N/A |
| SToRM | 25.8 ±12.0 | 0.20 ±0.04 † | 1.98 ±0.56+ | 2.12 ±0.55+ | 2.53 ±0.61 |
| VarNet | 55.1 ±35.7 †,* | 0.18 ±0.03 * | 2.82 ±0.42* | 2.78 ±0.41* | 2.36 ±0.55 |
| HyperSLICE | 42.5 ±20.3 †,* | 0.21 ±0.04 † | 1.20 ±0.44 *,+ | 1.10 ±0.3*,+ | 1.10 ±0.3*,+ |

**Table 3. Quantitative In Vivo Results: A) Acquisitions:** SNR, Edge Sharpness, Subjective Image Sharpness, Subjective Artefact and Subjective Motion (1- Worst, 5-Best) mean values and standard deviation computed over the prospective images (10 subjects) for Reference Breath-hold, Cartesian real-time and HyperSLICE reconstructed images. Values with '†' and '*' are statistically significantly different to Reference Breath-hold and Real-time Cartesian values respectively (post-hoc Nemenyi test, p<0.05). **B) Reconstructions:** SNR, Edge Sharpness, Subjective ranking of Image Sharpness, Artefact and Motion (1- Best, 3-Worst) mean values and standard deviation computed over the prospective images (10 subjects). Additionally includes Spiral Gridded SNR and Edge sharpness. Values with '†', '*' and '+' are statistically significantly different to Gridded, SToRM and VarNet values respectively (post-hoc Nemenyi test, p<0.05).

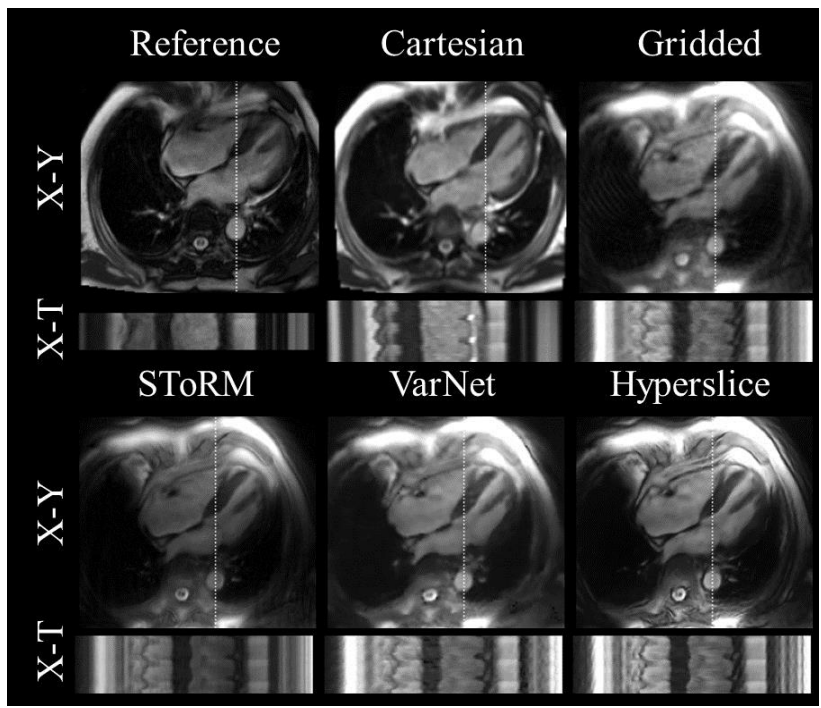

**Figure 5. Qualitative In-Vivo Results:** X-Y and X-T (white dotted line) views in the 4 chamber orientation seen with Reference breath-hold Cartesian, Cartesian real-time, optimized spiral trajectory with gridded reconstruction, optimized spiral trajectory with navigator-less SToRM (SToRM) reconstruction, optimized spiral trajectory with VarNet reconstruction and proposed HyperSLICE method.

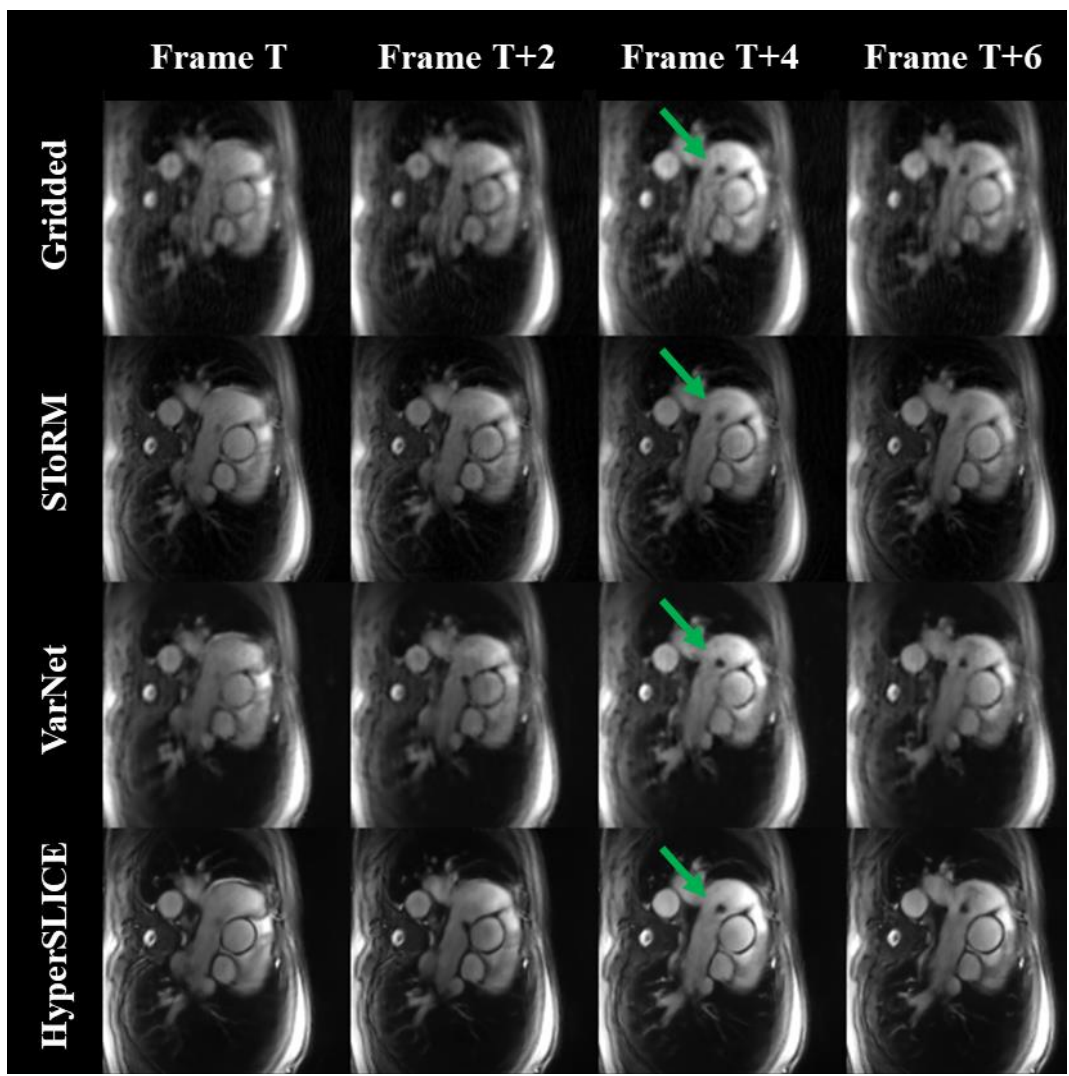

**Figure 6. Prospective Catheter Images.** Optimized spiral data of a sequence of images (every other frame shown). as reconstructed using Gridding, SToRM, VarNet and HyperSLICE. The catheter is indicated by the green arrow in the frame T+4. Only HyperSLICE and gridded reconstructions can be performed interactively. The corresponding video is provided in Supporting Information Video S3.

## Supporting Figures

| Acquisition Parameters | Type | Spatial resolution (mm²) | Temporal Resolution (ms) | TE/TR (ms) | Flip Angle (º) | Acceleration | Nominal FOV (mm²) |
|---|---|---|---|---|---|---|---|
| Reference Cartesian | Breath-hold | 1.5x1.5 | 27.3 | 1.27/3.0 | 50 | 2 | 370x370 |
| Real time Cartesian | Real time | 2.5x2.5 | 97 | 0.98/2.4 | 50 | 2 | 320x320 |
| Optimized Spiral | Real time | 1.7x1.7 | 55 | 0.9/3.67 | 70 | Variable density R=[1.1-15.0] | 400x400 |

**Supporting Information Table S1.** Nominal acquisition parameters for reference Cartesian breath-hold, real-time Cartesian and optimized spiral trajectories acquired prospectively in patients.

| Spiral Parameters | End of inner k-space | Inner acceleration rate | Start of outer k-space | Relative acceleration outer/inner | Transition | Ordering | Validation SSIM |
|---|---|---|---|---|---|---|---|
| Range investigated | [0.1:0.3] | [12:24] | [inner:1-inner] | [0.01:0.35] | [linear, Hanning, quadratic] | [linear, tiny golden angle] | N/A |
| Optimized #1 | 0.15 | 16 | 0.60 | 0.07 | hanning | linear | 0.858 |
| Optimized #2 | 0.16 | 12 | 0.40 | 0.05 | linear | linear | 0.856 |
| Optimized #3 | 0.11 | 16 | 0.96 | 0.05 | hanning | tiny | 0.856 |

**Supporting Information Table S2.** The top three variable density spiral trajectories obtained from the HyperBand optimization.

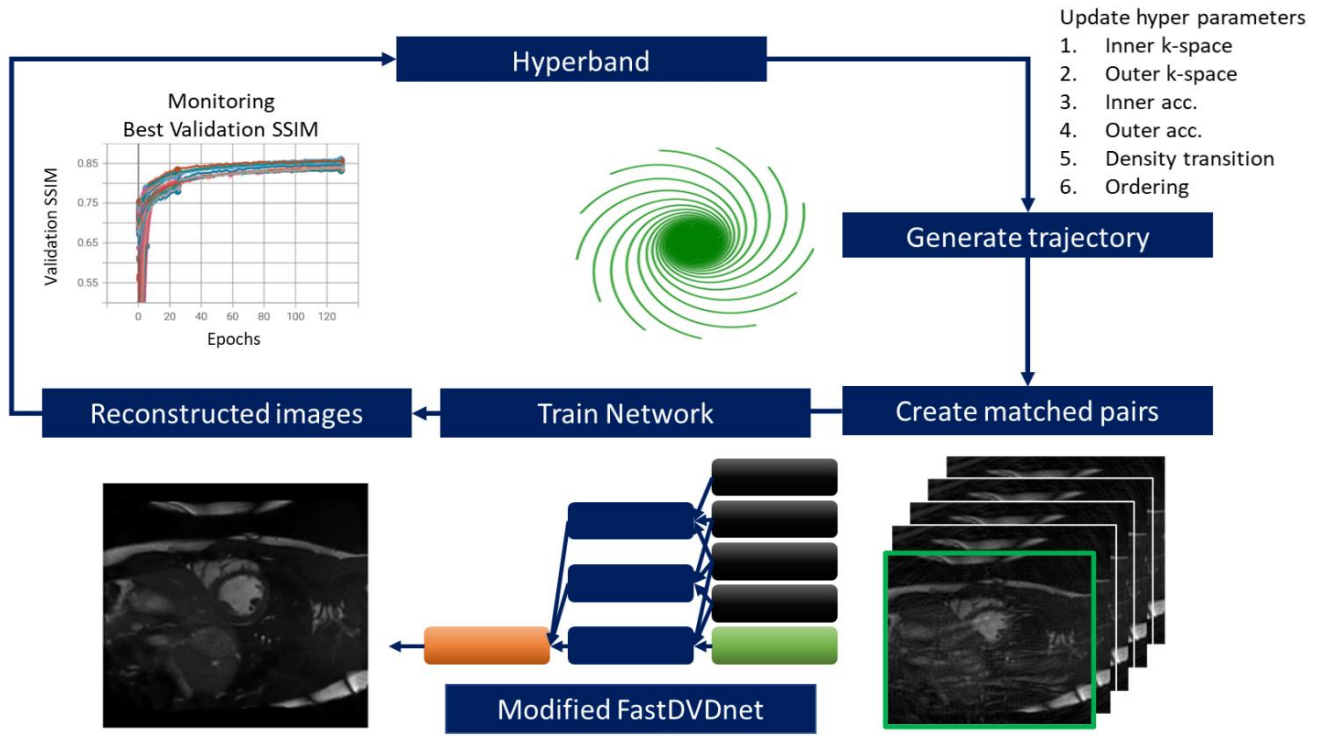

**Supporting Information Figure S1. Trajectory optimization through HyperBand:** Hyperparameters to generate spiral trajectories are updated depending on the resulting SSIM scores obtained from deep artifact suppressed images. The trajectory shown in green corresponds to the resulting optimized trajectory (parameters in Table 1).

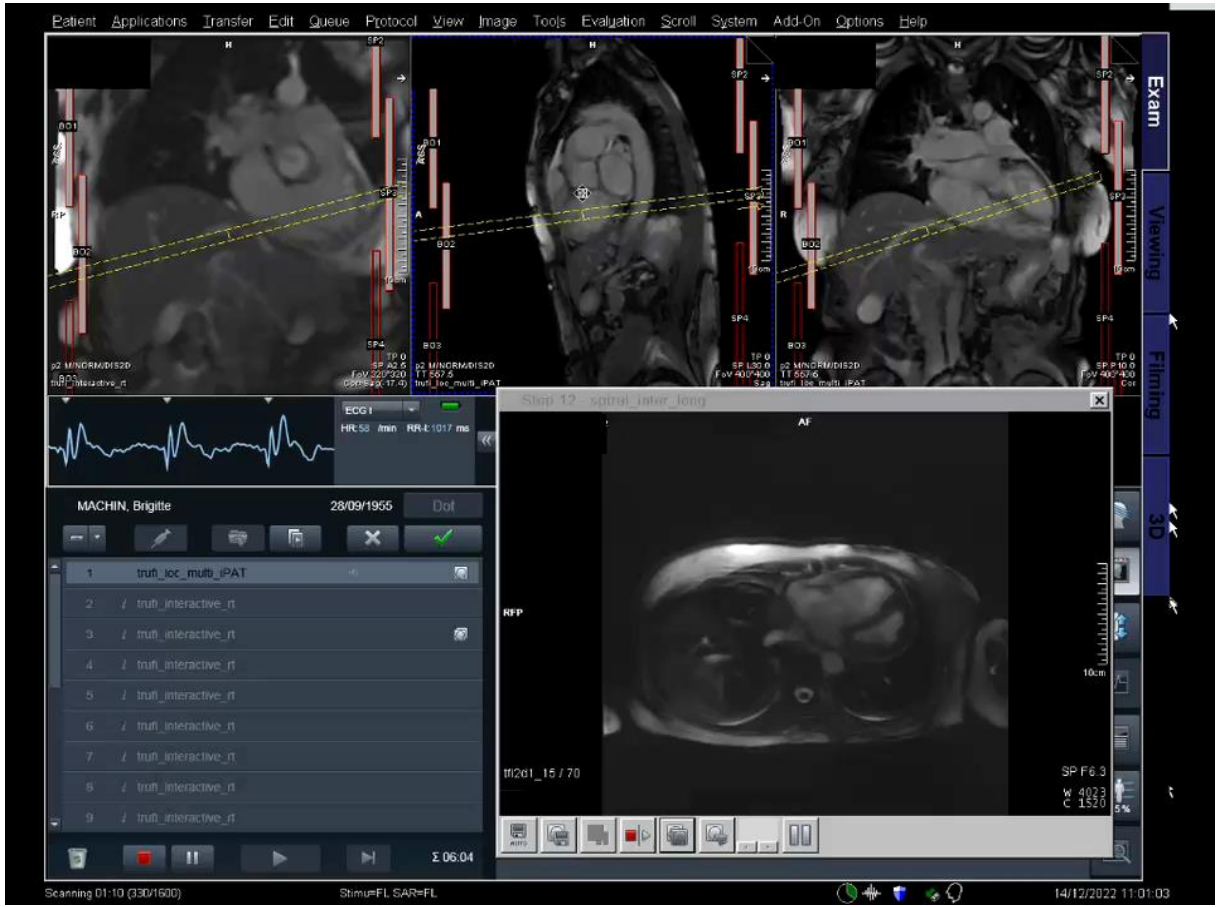

**Supporting Information Video S1.** Interactive imaging as performed on the scanner during catheter pull-back. The interface allows the user to move the scan plane and images update with minimal latency enabling immediate feedback on the location of the catheter.

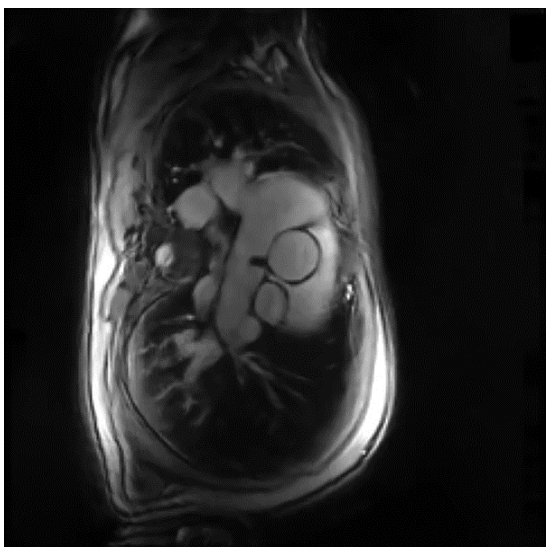

**Supporting Information Video S2.** Interactive imaging during catheter pull-back in a second patient with suspected pulmonary hypertension.

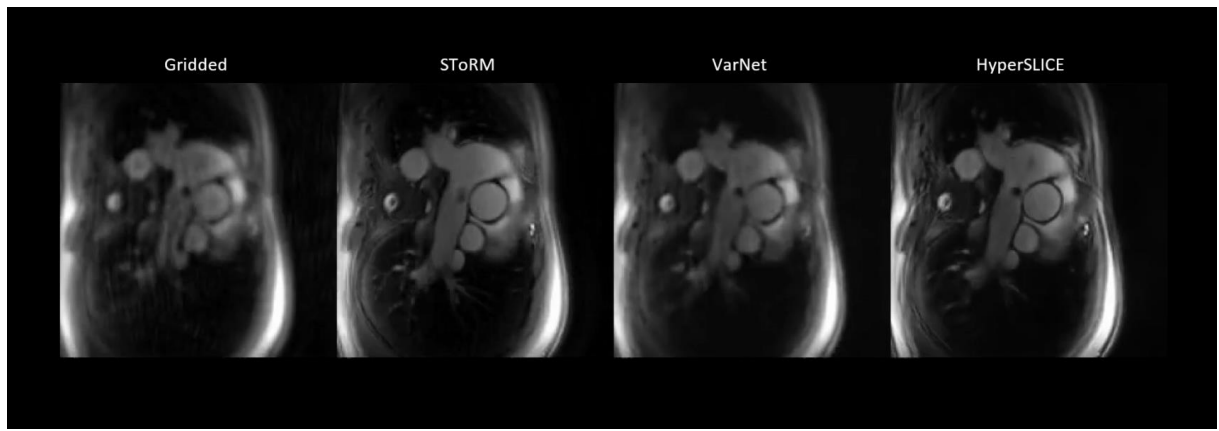

**Supporting Information Video S3.** Segment of the interactive scan during catheter pull-back reconstructed using all methods. The reconstructed segment had to be of a fixed orientation to compute coil sensitivities once for the comparison methods.

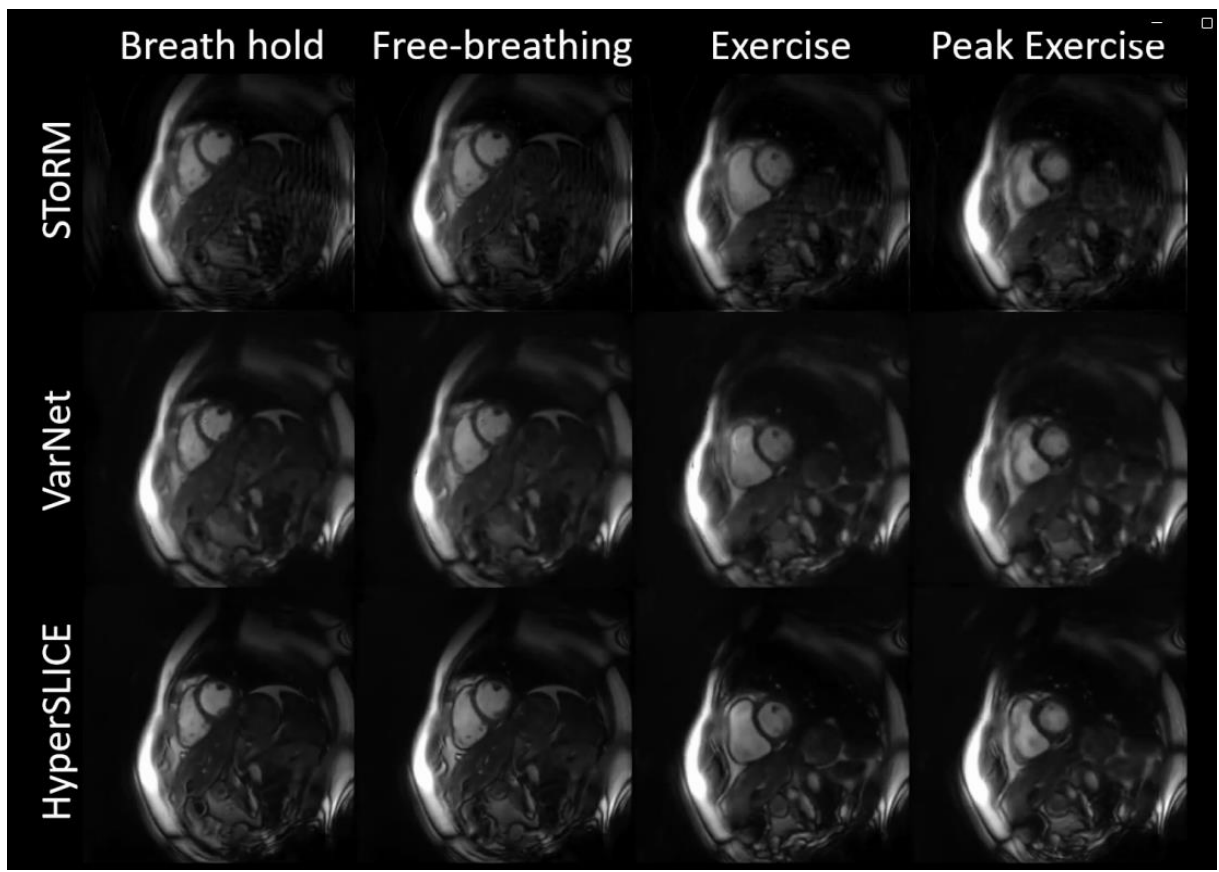

**Supporting Information Video S4.** SToRM, VarNet and HyperSLICE reconstructions of the same optimized spiral data acquired with different amounts of motion. From left to right: During breath-hold, free-breathing, exercise and peak exercise.

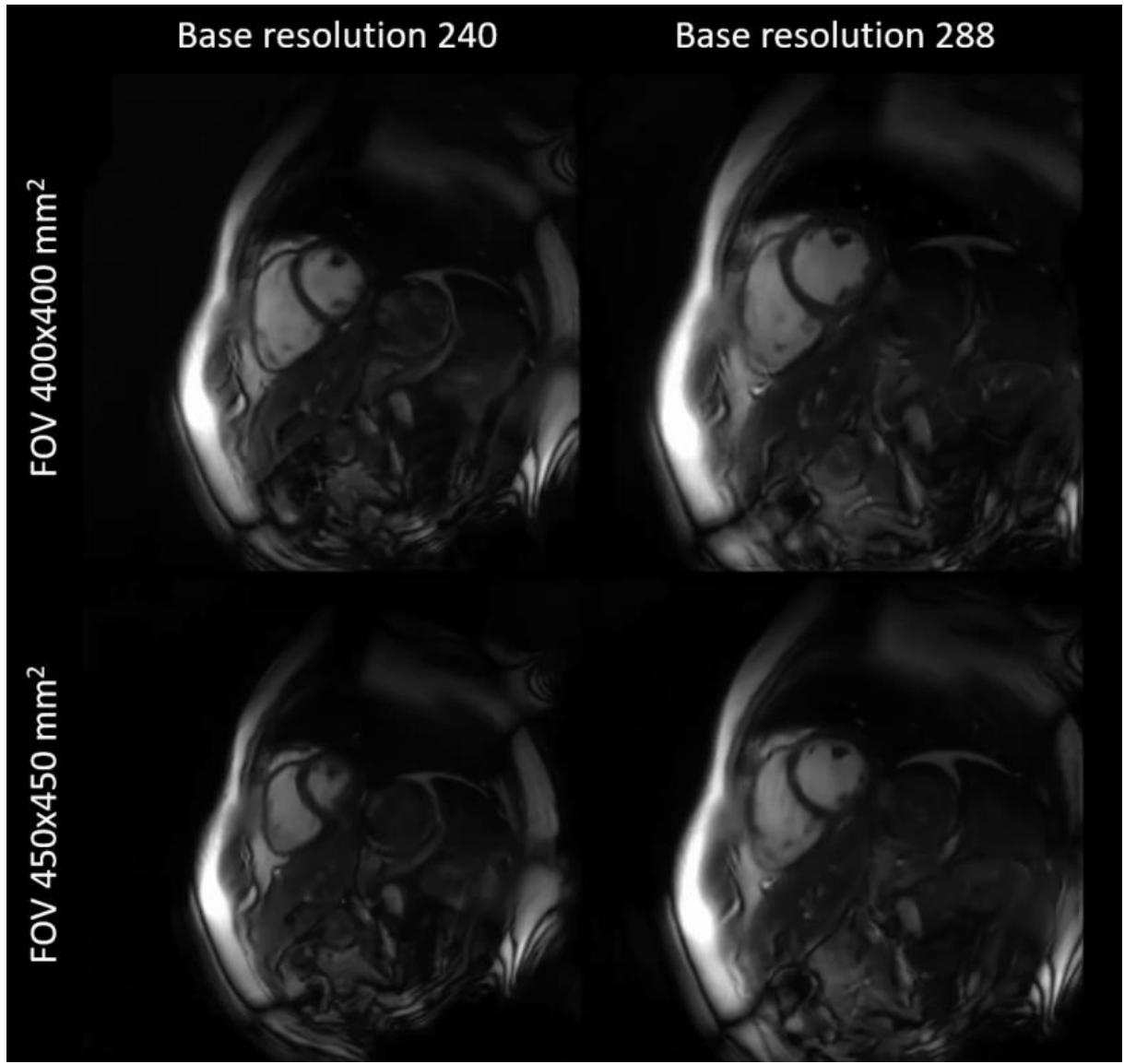

**Supporting Information Video S5.** HyperSLICE reconstructions using the same trained network of data acquired with: 1) the original FOV (400x400 mm$^2$) and base resolution (240), 2) higher base resolution (288), 3) larger FOV (FOV - 450x450mm$^2$) and 4) both changes (FOV - 450x450mm$^2$, base resolution - 288).

# References

1. Kerr AB, Pauly JM, Hu BS, et al. Real-time interactive MRI on a conventional scanner. *Magn Reson Med*. 1997;38(3):355-367. doi:10.1002/mrm.1910380303

2. Feng L, Srichai MB, Lim RP, et al. Highly accelerated real-time cardiac cine MRI using k–t SPARSE-SENSE. *Magn Reson Med*. 2013;70(1):64-74. doi:10.1002/MRM.24440

3. Lustig M, Donoho D, Pauly JM. Sparse MRI: The application of compressed sensing for rapid MR imaging. *Magn Reson Med*. 2007;58(6):1182-1195. doi:10.1002/mrm.21391

4. Schaetz S, Voit D, Frahm J, Uecker M. Accelerated Computing in Magnetic Resonance Imaging: Real-Time Imaging Using Nonlinear Inverse Reconstruction. *Comput Math Methods Med*. 2017;2017. doi:10.1155/2017/3527269

5. Hirsch FW, Frahm J, Sorge I, Roth C, Voit D, Gräfe D. Real-time magnetic resonance imaging in pediatric radiology — new approach to movement and moving children. *Pediatr Radiol*. 2021;51(5):840-846. doi:10.1007/S00247-020-04828-5/FIGURES/4

6. Unterberg-Buchwald C, Ritter CO, Reupke V, et al. Targeted endomyocardial biopsy guided by real-time cardiovascular magnetic resonance. *Journal of Cardiovascular Magnetic Resonance*. 2017;19(1):45. doi:10.1186/s12968-017-0357-3

7. Knoll F, Murrell T, Sriram A, et al. Advancing machine learning for MR image reconstruction with an open competition: Overview of the 2019 fastMRI challenge. *Magn Reson Med*. 2020;84(6):3054-3070. doi:10.1002/mrm.28338

8. Muckley MJ, Riemenschneider B, Radmanesh A, et al. State-of-the-Art Machine Learning MRI Reconstruction in 2020: Results of the Second fastMRI Challenge.

*ArXiv*. December 2020:2012.06318v2. http://arxiv.org/abs/2012.06318. Accessed January 28, 2021.

9. Hauptmann A, Arridge S, Lucka F, Muthurangu V, Steeden JA. Real-time cardiovascular MR with spatio-temporal artifact suppression using deep learning– proof of concept in congenital heart disease. *Magn Reson Med*. 2019;81(2):1143-1156. doi:10.1002/mrm.27480

10. Jaubert O, Montalt-Tordera J, Knight D, et al. Real-time deep artifact suppression using recurrent U-Nets for low-latency cardiac MRI. *Magn Reson Med*. 2021;86(4):1904-1916. doi:10.1002/MRM.28834

11. Li L, Jamieson K, DeSalvo G, Rostamizadeh A, Talwalkar A. Hyperband: A Novel Bandit-Based Approach to Hyperparameter Optimization. *Journal of Machine Learning Research*. 2016;18:1-52. https://arxiv.org/abs/1603.06560v4. Accessed February 24, 2022.

12. Tassano M, France G, Delon J, Veit T. FastDVDnet: Towards Real-Time Deep Video Denoising Without Flow Estimation. 2020:1354-1363.

13. Abadi M, Barham P, Chen J, et al. TensorFlow: A System for Large-Scale Machine Learning TensorFlow: A system for large-scale machine learning. *Proceedings of the 12th USENIX Symposium on Operating Systems Design and Implementation*. 2016:265-283. https://tensorflow.org. Accessed October 12, 2020.

14. Pipe JG, Zwart NR. Spiral trajectory design: A flexible numerical algorithm and base analytical equations. *Magn Reson Med*. 2014;71(1):278-285. doi:10.1002/MRM.24675

15. Winkelmann S, Schaeffter T, Koehler T, Eggers H, Doessel O. An Optimal Radial Profile Order Based on the Golden Ratio for Time-Resolved MRI. *IEEE Trans Med Imaging*. 2007;26(1):68-76. doi:10.1109/TMI.2006.885337

16. Bahnemiri SG, Ponomarenko M, Egiazarian K. On Verification of Blur and Sharpness Metrics for No-reference Image Visual Quality Assessment. *2020 IEEE 22nd International Workshop on Multimedia Signal Processing (MMSP)*. September 2020:1-6. doi:10.1109/MMSP48831.2020.9287110

17. Subbarao M, Choi TS, Nikzad A. Focusing techniques. *https://doi.org/101117/12147706*. 1993;32(11):2824-2836. doi:10.1117/12.147706

18. Hansen MS, Sørensen TS. Gadgetron: An open source framework for medical image reconstruction. *Magn Reson Med*. 2013;69(6):1768-1776. doi:10.1002/mrm.24389

19. Ahmed AH, Zhou R, Yang Y, Nagpal P, Salerno M, Jacob M. Free-Breathing and Ungated Dynamic MRI Using Navigator-Less Spiral SToRM. *IEEE Trans Med Imaging*. 2020;39(12):3933-3943. doi:10.1109/TMI.2020.3008329

20. Kleineisel J, Heidenreich JF, Eirich P, et al. Real-time cardiac MRI using an undersampled spiral k-space trajectory and a reconstruction based on a variational network. *Magn Reson Med*. 2022;88(5):2167-2178. doi:10.1002/mrm.29357

21. Montalt Tordera J. TensorFlow MRI. September 2022. doi:10.5281/ZENODO.7120930

22. Steeden JA, Atkinson D, Hansen MS, Taylor AM, Muthurangu V. Rapid flow assessment of congenital heart disease with high-spatiotemporal- resolution gated spiral phase-contrast MR imaging. *Radiology*. 2011;260(1):79-87. doi:10.1148/radiol.11101844

23. McGibney G, Smith MR. An unbiased signal-to-noise ratio measure for magnetic resonance images. *Med Phys*. 1993;20(4):1077-1078. doi:10.1118/1.597004

24. Virtanen P, Gommers R, Oliphant TE, et al. SciPy 1.0: fundamental algorithms for scientific computing in Python. *Nat Methods*. 2020;17(3):261-272. doi:10.1038/S41592-019-0686-2

25. Aggarwal HK, Jacob M. J-MoDL: Joint Model-Based Deep Learning for Optimized Sampling and Reconstruction. November 2019. doi:10.1109/JSTSP.2020.3004094

26. Wang G, Luo T, Nielsen JF, Noll DC, Fessler JA. B-Spline Parameterized Joint Optimization of Reconstruction and K-Space Trajectories (BJORK) for Accelerated 2D MRI. *IEEE Trans Med Imaging*. 2022;41(9):2318-2330. doi:10.1109/TMI.2022.3161875

27. Zhang J, Zhang H, Wang A, et al. Extending LOUPE for K-space Under-sampling Pattern Optimization in Multi-coil MRI. July 2020. doi:10.48550/arXiv.2007.14450

28. Cao J, Wang Q, Liang J, Zhang Y, Zhang K, van Gool L. Practical Real Video Denoising with Realistic Degradation Model. August 2022. doi:10.48550/arxiv.2208.11803

29. Vannesjo SJ, Haeberlin M, Kasper L, et al. Gradient system characterization by impulse response measurements with a dynamic field camera. *Magn Reson Med*. 2013;69(2):583-593. doi:10.1002/MRM.24263